\documentclass{article}

\usepackage[utf8]{inputenc}
\usepackage[T1]{fontenc}
\usepackage{microtype}
\usepackage{graphicx}
\usepackage{subcaption}
\usepackage{booktabs}
\usepackage{hyperref}

\usepackage[accepted]{icml2026}
\makeatletter
\renewcommand{\ICML@appearing}{\textit{Mechanistic Interpretability Workshop at
the 43rd International Conference on Machine Learning}, Seoul, South Korea,
2026. Copyright 2026 by the author(s).}
\makeatother
\hypersetup{pdfsubject={Mechanistic Interpretability Workshop at the 43rd
International Conference on Machine Learning, 2026}}

\usepackage{amsfonts}
\usepackage{amsmath}
\usepackage{amssymb}
\usepackage{nicefrac}
\usepackage{xcolor}
\usepackage{multirow}
\usepackage{array}

\graphicspath{{img/}}

\newcommand{\dmodel}{d_{\mathrm{model}}}
\newcommand{\dsae}{d_{\mathrm{sae}}}

\icmltitlerunning{Discovering Cross-Language Reasoning Invariance in LLMs with Geometry-Invariant Sparse Autoencoders}

\begin{document}

\twocolumn[
  \icmltitle{Discovering Cross-Language Reasoning Invariance in LLMs with
  Geometry-Invariant Sparse Autoencoders}

  \begin{icmlauthorlist}
    \icmlauthor{Igor Bogdanov}{carleton}
    \icmlauthor{Changcheng Huang}{carleton}
  \end{icmlauthorlist}

  \icmlaffiliation{carleton}{Department of Systems and Computer Engineering,
  Carleton University, Ottawa, ON, Canada}
  \icmlcorrespondingauthor{Igor Bogdanov}{igorbogdanov@cmail.carleton.ca}

  \icmlkeywords{Mechanistic Interpretability, Multilingual Large Language Models,
Sparse Autoencoders, Cross-Lingual Representations, Multilingual Reasoning,
Causal Patching}

  \vskip 0.3in
]

\printAffiliationsAndNotice{}

\begin{abstract}
% ---- inlined from parts/main/00_abstract ----
Multilingual language models can solve the same mathematical reasoning problem in different languages, but it remains unclear whether they rely on shared internal features or on language-specific computations that only produce similar outputs. We study this question in five models from four architecture families using the Multilingual Grade School Math (MGSM) dataset, with problems solved independently in English, German, French, Spanish, Russian, and Chinese, retaining only problems with valid reasoning traces in all six languages and replaying those traces through the model to record internal representations at multiple layers. For each model, we first use Centered Kernel Alignment (CKA) to identify layers with strong cross-language alignment. At each selected layer, we train two sparse autoencoders: a baseline reconstruction-only model and a contrastive variant introduced in this work, the Geometry-Invariant Sparse Autoencoder (GI-SAE). GI-SAE supplements the reconstruction loss with an Information Noise-Contrastive Estimation (InfoNCE) loss that trains the encoder to produce similar feature activations for traces of the same problem, regardless of language or token position. We then test whether the resulting shared features are functionally interchangeable by swapping shared feature values between languages during the model's forward pass and measuring the resulting change in output (causal patching), quantified by Kullback-Leibler (KL) divergence per shared feature. Although GI-SAE yields higher CKA and Jaccard similarity at nearly every layer, higher geometric similarity does not consistently imply greater functional interchangeability across languages. We find that cross-language feature sharing is strongly model- and architecture-dependent in this sample and appears at different depths in different models. GI-SAE primarily amplifies cross-language structure already present in each model: the pattern is model-specific, with progressive strengthening in Qwen, no functional benefit in already-saturated Gemma, and mixed layer-dependent effects in Llama and Phi.
% ---- end parts/main/00_abstract ----
\end{abstract}

% ---- inlined from parts/main/main ----
% ---- inlined from parts/main/01_introduction ----
\section{Introduction}
\label{sec:introduction}

Multilingual large language models solve mathematical reasoning problems
across languages, but whether they develop \emph{shared} internal
representations for reasoning, or merely produce correct answers via
language-specific computations, remains an open question.
The distinction has practical consequences: shared representations
would allow a single interpretability analysis to cover all languages,
while language-specific computations would require separate \mbox{analysis} for
each language.

Three gaps remain in this literature:
to our knowledge, no study has tested whether cross-language shared
features are \emph{functionally interchangeable} (substituting values
preserves behavior); sharing has not been compared across model families
under a common protocol; and it is unknown whether geometric metrics
(CKA, Jaccard) reliably predict functional interchangeability.

We address these gaps by training two top-$K$
SAEs~\citep{gao2025scaling} at each layer of each model:
a \emph{baseline} SAE (reconstruction only) and a
\emph{geometry-invariant} SAE
(GI-SAE)~\citep{oord2018representation} that adds an InfoNCE
contrastive loss encouraging similar feature activations across
languages for the same problem. We test functional interchangeability
via causal patching: swapping shared feature values between languages
and measuring KL divergence per shared feature.

Across five models from four families (78 layer observations, two SAEs
each), we find that GI-SAE improves geometric similarity nearly
everywhere but improves functional interchangeability only where
baseline sharing is moderate. Geometric similarity alone does not
guarantee functional interchangeability.

We contribute a causal measurement framework for testing whether
cross-language SAE features are functionally interchangeable; GI-SAE,
a top-$K$ SAE with an InfoNCE objective; and a five-model,
six-language empirical study showing that baseline shared fraction
stratifies GI-SAE outcomes: 83\% win rate in the convergent profile
(95\% CI [64, 93]\%), no systematic benefit in the low-sharing profile,
and 6\% in the saturated profile.
% ---- end parts/main/01_introduction ----
% ---- inlined from parts/main/02_background ----
\section{Related Work}
\label{sec:related}

\paragraph{Multilingual representations in LLMs.}
Language-specific neurons can steer output
language~\citep{tang2024language, kojima2024multilingual}, and
multilingual transformers appear to route non-English inputs through a
shared latent space~\citep{wendler2024llamas, tezuka2025transfer}.
Cross-lingual reasoning ability has been linked to neuron
overlap~\citep{hu2025crosslingual}, though high behavioral consistency
does not necessarily imply shared internal
representations~\citep{ifergan2025beneath}.
Our work studies \emph{reasoning traces} rather than factual recall,
and evaluates sharing at the level of individual SAE features with
causal interventions across multiple checkpoints.

\paragraph{Sparse autoencoders for interpretability.}
SAEs decompose activations into sparse, interpretable
features~\citep{bricken2023monosemanticity, cunningham2024sparse};
\citet{gao2025scaling} characterize top-$K$ SAE scaling.
In multilingual settings, \citet{deng2025unveiling} identify
language-specific features and \citet{brinkmann2025shared} show that
SAE features encode shared grammatical concepts with causal validation.
We ask a different question: whether cross-language shared features are
functionally interchangeable, not just geometrically similar.

\paragraph{Causal methods for evaluating representations.}
Causal intervention tests functional
necessity~\citep{meng2022locating, conmy2023automated}. Our protocol
operates at individual SAE features. This matters because CKA can
assign high similarity to functionally different
representations~\citep{davari2023reliability}; we show that geometric
convergence~\citep{huh2024platonic} does not imply functional
interchangeability.
% ---- end parts/main/02_background ----
% ---- inlined from parts/main/03_method ----
\section{Methodology and Measurement Framework}
\label{sec:framework}

Each model solves the same math problems independently in six languages,
producing a reasoning trace per (problem, language) pair. We record the
model's internal state at selected layers during each trace, then train
two SAEs on these recorded activations: a baseline SAE trained only to
reconstruct them, whose learned features we compare across languages to
measure naturally occurring sharing, and GI-SAE, which adds a
contrastive loss that trains the encoder to produce similar feature
activations across languages for the same problem. Comparing the two
under causal patching reveals where cross-language shared features are
functionally interchangeable and where they are not.

\subsection{Layer Selection via CKA Survey}
\label{sec:framework-layers}

In a transformer, each layer adds its output to a running sum called
the \emph{residual stream}~\citep{elhage2021mathematical}. At layer $l$
and token position $t$, the residual-stream vector
$x^{(l)}_t \in \mathbb{R}^{\dmodel}$ is the cumulative representation
that all subsequent layers read from and write to.
All SAE training and causal interventions in this work operate on
residual-stream vectors. Patching the residual stream at layer~$l$
modifies the state read by all downstream layers $l{+}1, \ldots, L$.

Not all layers are equally relevant. To identify informative layers,
we compute linear CKA~\citep{kornblith2019similarity} between
pre-answer residual-stream matrices for each language pair at each
layer, with rows matched by problem ID (both matrices always from the
same layer). We define the \textbf{pre-answer token} as the final token
of the extracted reasoning span, immediately before the JSON answer
block; the final JSON block and its markup are excluded.
This token is used only for the CKA survey; later evaluations use
multiple backward offsets.
We select the contiguous depth range with highest mean pairwise CKA;
this range differs across models (Table~\ref{tab:models}) because
cross-language alignment peaks at different depths.

\subsection{Activation Extraction and Normalization}
\label{sec:framework-extraction}

At identified layers, we replay each reasoning trace through the model
using TransformerLens~\citep{nanda2022transformerlens} and record the
residual-stream vector at every token position, producing one
$T \times \dmodel$ matrix per (problem, language, layer) triple.

\paragraph{Normalization and patching convention.}
SAE training and geometric evaluation use z-scored activations
(per-coordinate statistics from the training split). The main
causal-patching sweep uses native-scale encoding; a normalized-space
sensitivity check on representative layers shows the
Qwen/convergent results are stable
(Appendix~\ref{app:norm-sensitivity}).

\subsection{Sparse Autoencoder Architecture}
\label{sec:framework-sae}

We use a top-$K$ sparse autoencoder~\citep{gao2025scaling} with encoder
$W_\text{enc} \in \mathbb{R}^{\dsae \times \dmodel}$, decoder
$W_\text{dec} \in \mathbb{R}^{\dmodel \times \dsae}$, and biases.
Given a single input activation $x \in \mathbb{R}^{\dmodel}$
(one residual-stream vector at one token position and one layer):
\begin{equation}
  \begin{aligned}
    z &= W_\text{enc}\, x + b_\text{enc},\;
    f = \operatorname{ReLU}\!\bigl(\operatorname{TopK}(z, K)\bigr), \\
    \hat{x} &= W_\text{dec}\, f + b_\text{dec}.
  \end{aligned}
  \label{eq:sae}
\end{equation}
Sparsity is enforced by the top-$K$ bottleneck ($K{=}128$); no
$\ell_1$ penalty is used. The expansion factor $4{\times}$
($\dsae = 4 \dmodel$) controls dictionary size (8{,}192--12{,}288
features; Table~\ref{tab:models}), while $K$ controls per-input
sparsity. Decoder columns are unit-$\ell_2$ normalized after each step.
The \textbf{baseline SAE} is trained with MSE reconstruction loss
$\mathcal{L}_\text{recon}$; by comparing which features activate
across languages under this objective, we observe naturally occurring
cross-language sharing.

\subsection{Geometry-Invariant SAE (GI-SAE)}
\label{sec:framework-gisae}

GI-SAE uses the same encoder-decoder architecture
(Equation~\ref{eq:sae}) but adds an Information
Noise-Contrastive Estimation
(InfoNCE)~\citep{oord2018representation} contrastive term to the
reconstruction loss.
Each training activation is labeled with its source problem ID
$y_i \in \{1, \ldots, Q\}$, where $Q$ is the number of training
problems. Activations sharing the same problem ID, regardless of
language or token position, form \emph{positive pairs}; activations
from different problems are \emph{negatives}:
\begin{equation}
  \mathcal{L}_\text{contr} = -\frac{1}{|B'|}\sum_{i \in B'} \log
    \!\left(
    \frac{\sum_{j \in P_i} \exp(s_{ij})}
         {\sum_{k \neq i} \exp(s_{ik})}
    + \epsilon
    \right)\!,
  \label{eq:infonce}
\end{equation}
where $s_{ij} = \tilde{f}_i \cdot \tilde{f}_j / \tau$ is the cosine similarity of
$\ell_2$-normalized features divided by temperature $\tau{=}0.1$,
$P_i = \{j : y_j = y_i,\, j \neq i\}$ is the positive set,
$B' = \{i : |P_i| > 0\}$ excludes items without positives,
and $\epsilon = 10^{-8}$ prevents $\log 0$ when no positive pairs exist
in the batch. Self-pairs are excluded from the denominator.

The joint objective is:
\begin{equation}
  \mathcal{L}_\text{GI}
  = \mathcal{L}_\text{recon} + w \cdot \mathcal{L}_\text{contr}.
  \label{eq:gi-loss}
\end{equation}
We use $w{=}1.0$, selected on Qwen3-1.7B L20 and fixed across all
models (Appendix~\ref{app:training}).

\subsection{Backward Alignment of Reasoning Trajectories}
\label{sec:framework-backward}

The same reasoning step may occur at different absolute token positions
across languages. We therefore anchor at the last reasoning token and
measure \textbf{backward offsets} ($-1$ to $-500$). The contrastive
dataset labels all tokens by problem ID; positives include
same-language activations from other positions, so cross-language
invariance is tested explicitly by Jaccard similarity and causal
patching rather than imposed solely by the objective.
This targets the answer-proximal phase of the trace, where discourse
and answer-transition structure are most comparable across languages.

\subsection{Causal Validation via Feature Patching}
\label{sec:framework-patching}

Geometric metrics measure whether the same features activate but not
whether swapping their values preserves behavior. We validate with
\textbf{causal patching} (Algorithm~\ref{alg:patch}).

\begin{algorithm}[h!]
\caption{Causal feature patching}\label{alg:patch}
\begin{algorithmic}[1]
\REQUIRE Target residual $r_\text{tgt}$, donor residual $r_\text{src}$, trained SAE
\ENSURE KL divergence, autoregressive flip indicator
\STATE $f_\text{tgt} \gets \operatorname{encode}(r_\text{tgt})$;\;
       $f_\text{src} \gets \operatorname{encode}(r_\text{src})$
\STATE $S \gets \{j : f_{\text{tgt},j} > 0 \;\wedge\; f_{\text{src},j} > 0\}$
  \COMMENT{shared active features}
\STATE $f_\text{patched} \gets f_\text{tgt}$
\STATE $f_\text{patched}[S] \gets f_\text{src}[S]$
  \COMMENT{swap shared values from donor}
\STATE $r_\text{patched} \gets \operatorname{decode}(f_\text{patched})
  + \bigl(r_\text{tgt} - \operatorname{decode}(f_\text{tgt})\bigr)$
  \COMMENT{preserve reconstruction error}
\STATE Run forward pass with $r_\text{patched}$ replacing $r_\text{tgt}$
\STATE \textbf{return} $\mathrm{KL}(p_\text{clean} \| p_\text{patched})$,\;
  $\mathbf{1}[\arg\max p_\text{patched} \neq \arg\max p_\text{clean}]$
  \COMMENT{disruption, flip}
\end{algorithmic}
\end{algorithm}

Each trace is replayed using its original token sequence; the main sweep uses
native-scale patching (Appendix~\ref{app:norm-sensitivity} compares
to normalized-space patching). The reconstruction error is preserved
(line~5), so the intervention modifies only the SAE feature subspace.
Output disruption is measured by KL
divergence~\citep{kullback1951information} between the clean and
patched next-token distributions at the final reasoning-token position.
We also record an \emph{autoregressive flip} (top-1 prediction change).
Same-language pairs serve as near-zero disruption controls.
Full implementation details are in
Appendix~\ref{app:patching-details}.

\subsection{Evaluation Metrics and Decision Rule}
\label{sec:framework-metrics}

\paragraph{Geometric metrics (representation-level).}
\textbf{Linear CKA}~\citep{kornblith2019similarity} measures geometric
similarity between SAE-encoded activation matrices of two languages for
the same problem.
\textbf{Jaccard similarity}~\citep{manning2008introduction} at backward
offset $\delta$: the set overlap of active feature indices, defined as
$J = |S_A \cap S_B| / |S_A \cup S_B|$.

\paragraph{Causal metric (function-level).}
We define the \textbf{shared fraction} as the proportion of $K$ features active
in both languages: $\phi = \overline{|S|}_\text{cross} / K$.
The \textbf{autoregressive flip rate} (AFR) is the fraction of patched
traces where the model's top-1 next-token prediction at the pre-answer
position changes:
$\text{AFR} = \frac{1}{N}\sum_i \mathbf{1}[\arg\max p_{\text{patched},i}
  \neq \arg\max p_{\text{clean},i}]$.
AFR measures local disruption to the reasoning-to-answer transition; it
does not require continuing generation or parsing the final numerical
answer.
The primary metric, \textbf{KL/shared feature}, normalizes the cross-language
disruption by the number of features intervened on, corrected for the
same-language baseline:
\begin{equation}
  \text{KL/feat} = \frac{\overline{\text{KL}}_\text{cross} - \overline{\text{KL}}_\text{same}}
                        {\overline{|S|}_\text{cross}}.
  \label{eq:kl-per-feat}
\end{equation}
where $\overline{\text{KL}}_\text{cross}$,
$\overline{\text{KL}}_\text{same}$, and
$\overline{|S|}_\text{cross}$ denote means over all (problem,
language pair, backward offset) evaluation instances at that layer.
Lower values indicate greater functional interchangeability.

\paragraph{Decision rule.}
GI-SAE \emph{wins} at a layer when \\
$\text{KL/feat}_\text{GI} < \text{KL/feat}_\text{baseline}$.

\paragraph{Key distinction.}
CKA compares matrix geometry (rotation-invariant); Jaccard and shared
fraction compare active dictionary indices; causal patching tests
whether swapping shared-index values preserves behavior. These three
levels need not agree.
% ---- end parts/main/03_method ----
% ---- inlined from parts/main/04_experimental_setup ----
\section{Experimental Setup}
\label{sec:setup}

\paragraph{Models, families, and layers}
\label{sec:setup-models}

We evaluate five models from four architecture families, spanning 1.7B
to 4B parameters:
Qwen3-1.7B and Qwen3-4B~\citep{qwen2025qwen3},
Llama-3.2-3B~\citep{grattafiori2024llama},
Gemma-3-4B~\citep{team2025gemma}, and
Phi-3-mini-4k~\citep{abdin2024phi}.
Table~\ref{tab:models} summarizes each model's architecture and sweep
range. The two Qwen models constitute a within-family, same-generation,
different-scale comparison; Gemma, Llama, and Phi each represent a
distinct family.

\begin{table*}[t]
  \caption{Models, layer sweeps, and dataset sizes. $\dsae = 4 \times \dmodel$;
    $K{=}128$ for all models. Sweep ranges are determined by the CKA
    survey. Valid problems are those
    solved correctly in all six languages. Activation vectors are the
    total number of $\dmodel$-dimensional residual-stream vectors
    extracted across all tokens, languages, and swept layers.}
  \label{tab:models}
  \centering
  \footnotesize
  \begin{tabular}{llrrrrrr}
    \toprule
    Model & Family & Params & $\dmodel$ & Sweep & Valid & \multicolumn{2}{c}{Activation vectors} \\
    \cmidrule(l){7-8}
     &  &  &  &  & problems & Train & Val \\
    \midrule
    Qwen3-1.7B   & Qwen  & 1.7B & 2048 & L16--L26 (11) & 151 & 2.77M & 0.65M \\
    Qwen3-4B     & Qwen  & 4B   & 2560 & L21--L34 (14) & 202 & 5.42M & 1.06M \\
    Llama-3.2-3B & Llama & 3B   & 3072 & L11--L27 (17) &  60 & 0.79M & 0.20M \\
    Gemma-3-4B   & Gemma & 4B   & 2560 & L15--L30 (16) & 137 & 1.95M & 0.39M \\
    Phi-3-mini   & Phi   & 3.8B & 3072 & L1--L20\phantom{0} (20) &  63 & 1.18M & 0.24M \\
    \midrule
    \multicolumn{5}{l}{Total} & 613 & 12.10M & 2.54M \\
    \bottomrule
  \end{tabular}
\end{table*}

Sweep ranges differ across models because they target the region of
high cross-language CKA identified by the preliminary survey.
Phi-3's sweep starts at L1 because its CKA peaks at L9 (${\sim}31\%$ depth), unlike all
other models which peak at 88--100\% depth.

\paragraph{Common 6-language reasoning protocol}
\label{sec:setup-languages}

We use Multilingual Grade School Math
(MGSM)~\citep{shi2022language} as the evaluation corpus: 250
grade-school math problems, each translated into multiple languages. Each model solves every problem
independently in six languages (en, de, fr, es, ru, zh), producing complete
chain-of-thought reasoning traces via nucleus
sampling~\citep{holtzman2020curious}
(temperature 0.6, top-$p$ 0.95, top-$k$ 20, max 2048 new tokens,
seed $42 + \text{problem index}$ for reproducibility).

A problem is retained as ``valid'' only if the model solves it correctly
\emph{and} produces an extractable reasoning trace in all six languages.
All results in this paper are therefore conditional on successful
multilingual reasoning; they should not be interpreted as estimating
feature sharing over the full MGSM distribution.
This yields 15 cross-language pairs and 6 same-language control pairs
per problem. Valid problem counts differ substantially across models
because multilingual mathematical competence varies
(Table~\ref{tab:models}).

\paragraph{Corpus, training, and pipeline}
\label{sec:setup-corpus}

Each problem is presented with a language-specific system prompt
instructing step-by-step reasoning and a JSON-formatted answer.
Reasoning traces are extracted from think tags (Qwen) or from text
preceding the final JSON answer block (Gemma, Llama, Phi); the
extracted text contains no JSON block or markup, though
natural-language answer statements may remain in the reasoning span
(full prompts and extraction rules in Appendix~\ref{app:prompts}).
Valid problems are split into train/validation/test
(70/15/15\%, seed 42). Both SAE variants use Adam with learning rate
$10^{-4}$, batch size 256, maximum 200 epochs, and early stopping
(patience 30, monitoring validation MSE). All experiments run on a
single NVIDIA RTX 5090 GPU. The per-model pipeline has six stages
(activation extraction, SAE training, geometric evaluation, causal
patching, aggregation, analysis); each model takes 7--10 hours, with
the full sweep requiring ${\sim}$60 GPU-hours
(Appendix~\ref{app:patching-details}).
% ---- end parts/main/04_experimental_setup ----
% ---- inlined from parts/main/05_results ----
\section{Results}
\label{sec:results}

\begin{figure*}[t]
  \centering
  \includegraphics[width=\textwidth]{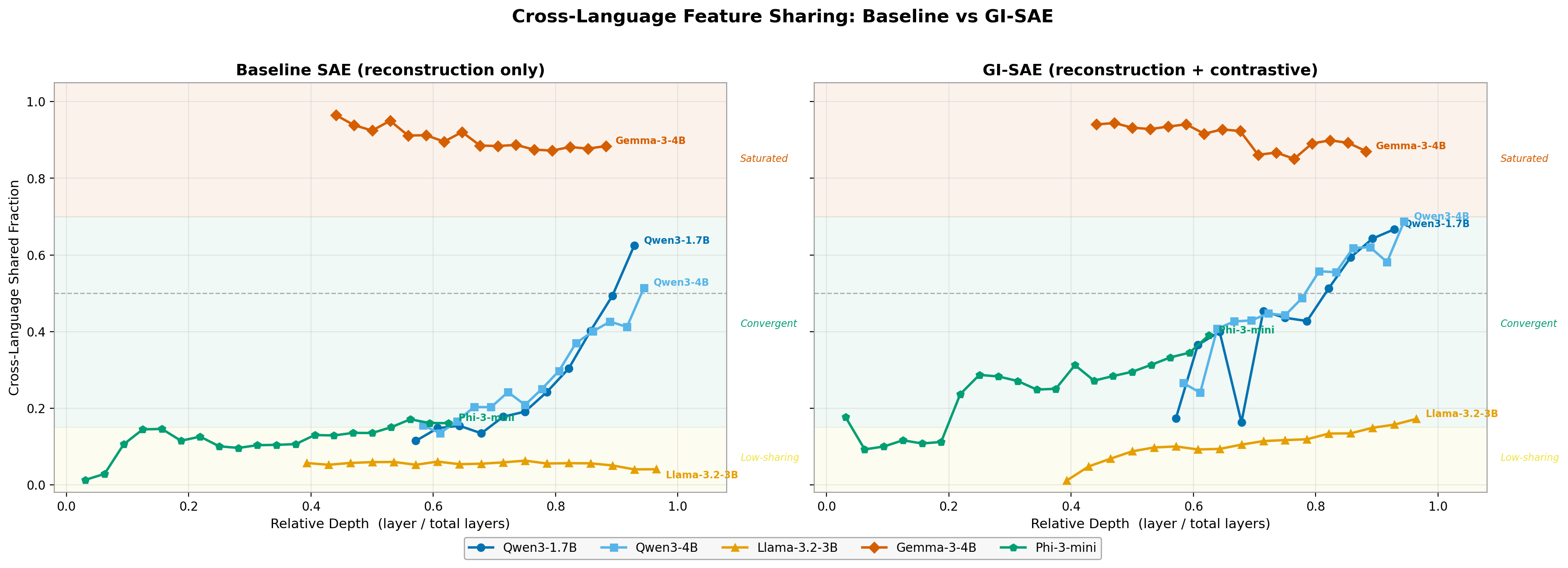}
  \caption{Cross-language shared fraction (fraction of 128 active features
    shared between two languages for the same problem) vs.\ relative depth.
    \textbf{Left}: baseline SAE (reconstruction only).
    \textbf{Right}: GI-SAE (reconstruction + contrastive).
    Colored bands mark the three sharing profiles:
    low-sharing ($<$15\%, orange), convergent (15--60\%, green),
    saturated ($>$60\%, red).}
  \label{fig:profiles}
\end{figure*}

\paragraph{Cross-language feature sharing: baseline and GI-SAE}
The baseline SAE is trained only to reconstruct activations; it has no
cross-language objective. By comparing which of its learned features
activate for the same problem across languages, we observe three
baseline-sharing profiles and several model-specific depth trajectories
(Figure~\ref{fig:profiles}, left).

Gemma shares 87--96\% of features across all layers; Qwen rises from
11--15\% to 50--62\% with depth (both Qwen models track closely); Phi
and Llama remain at 1--17\% and 4--6\% respectively. Models of similar
parameter count occupy different profiles, suggesting architecture
matters more than scale. GI-SAE (Figure~\ref{fig:profiles}, right)
increases Qwen's sharing to 16--69\% and Phi's to 9--39\%, but leaves
Gemma unchanged and Llama only marginally improved.

\subsection{Geometric Similarity Does Not Imply Functional Interchangeability}
\label{sec:results-dissociation}

\begin{figure*}[t]
  \centering
  \includegraphics[width=0.85\linewidth]{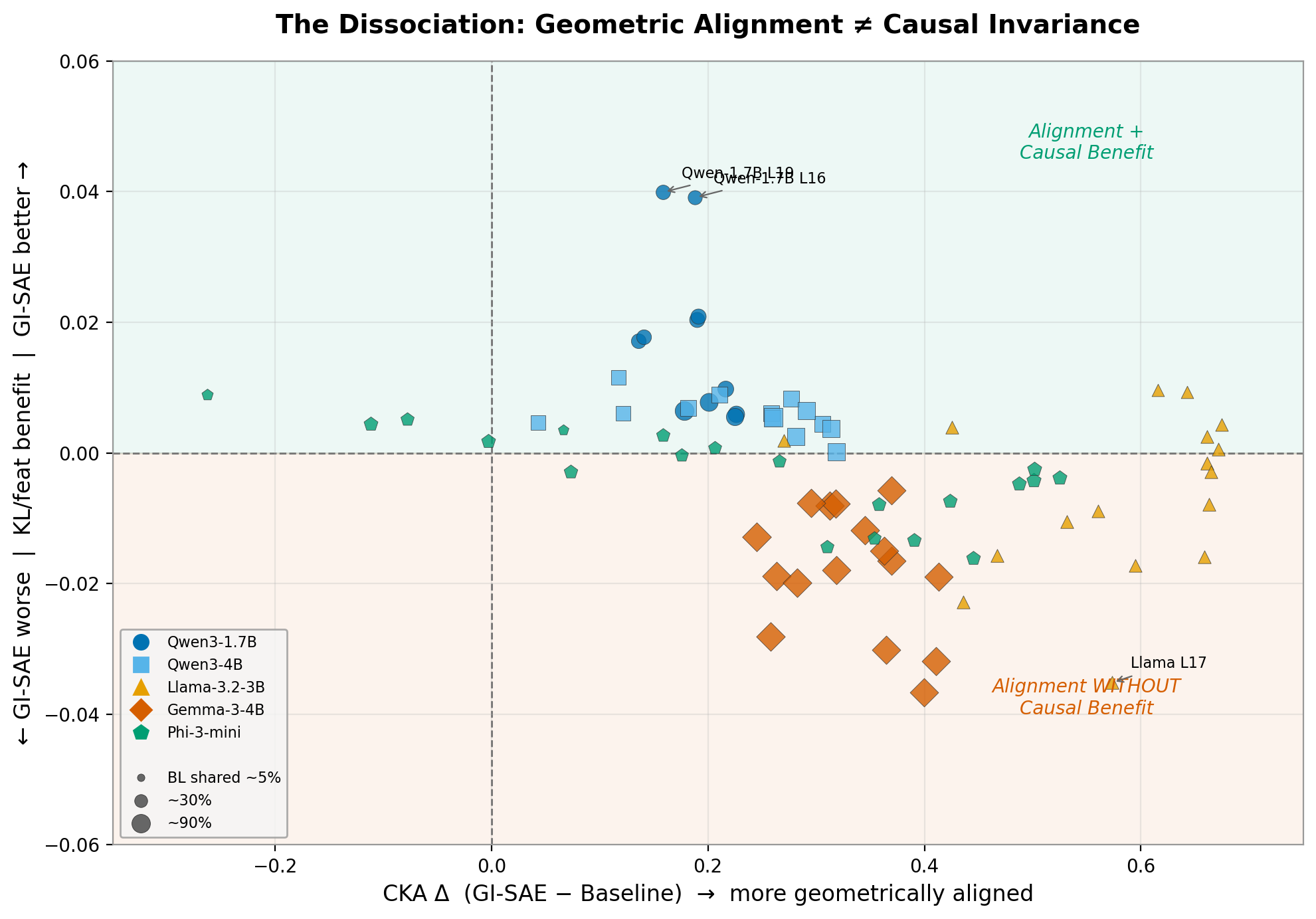}
  \caption{Each point is one (model, layer) observation.
    \textit{x}-axis: CKA improvement from GI-SAE over baseline
    (geometric similarity gain).
    \textit{y}-axis: KL/feature improvement (positive = GI-SAE has lower
    KL/feature).
    Marker size scales with baseline shared fraction.
    The upper-right quadrant (green) indicates both geometric and
    functional benefit; the lower-right (orange) indicates geometric
    improvement \emph{without} functional benefit.}
  \label{fig:dissociation}
\end{figure*}

GI-SAE improves CKA ($\Delta > 0$) at nearly every observation across all
five models (Figure~\ref{fig:dissociation}).
However, functional benefit does not follow uniformly:
\begin{itemize}
  \item \textbf{Qwen} occupies the upper-right quadrant: geometric similarity
    \emph{and} functional interchangeability improve together.
    CKA $\Delta$ +0.14--0.23 (1.7B) and +0.04--0.32 (4B); GI-SAE
    wins on 11/11 and 14/14 test layers, respectively.
  \item \textbf{Gemma} sits in the lower-right quadrant: CKA improves by
    +0.25--0.41, yet GI-SAE wins on 0/16 test layers.
  \item \textbf{Llama} achieves the largest CKA $\Delta$ in the study
    (up to +0.67) but wins on only 7/17 test layers; sustained
    functional gains appear in the deepest layers (L23--L27), with two
    isolated early wins (L12--L13).
  \item \textbf{Phi} is mixed: early layers (L1--L6) in the upper-right
    quadrant, mid-to-late layers in the lower-right.
\end{itemize}
This demonstrates that CKA improvement is \emph{not sufficient}
for functional interchangeability.
The gap between geometric similarity and functional interchangeability
is widest for models with low baseline sharing (Llama, Phi mid-layers)
and for the saturated model (Gemma).

Causal controls support the functional-interchangeability
interpretation: same-problem cross-language patches are less disruptive than
both different-problem donors (86\% pass rate) and random-value donors
(93\% pass rate), yielding an overall 89\% pass rate across 28
condition--variant pairs (Appendix~\ref{app:controls}). GI-SAE features
show stronger problem-specificity than baseline features (100\% vs.\
71\% on the different-problem control).

\subsection{A Phase Diagram for GI-SAE Success}
\label{sec:results-phase-diagram}

\begin{figure*}[t]
  \centering
  \includegraphics[width=\textwidth]{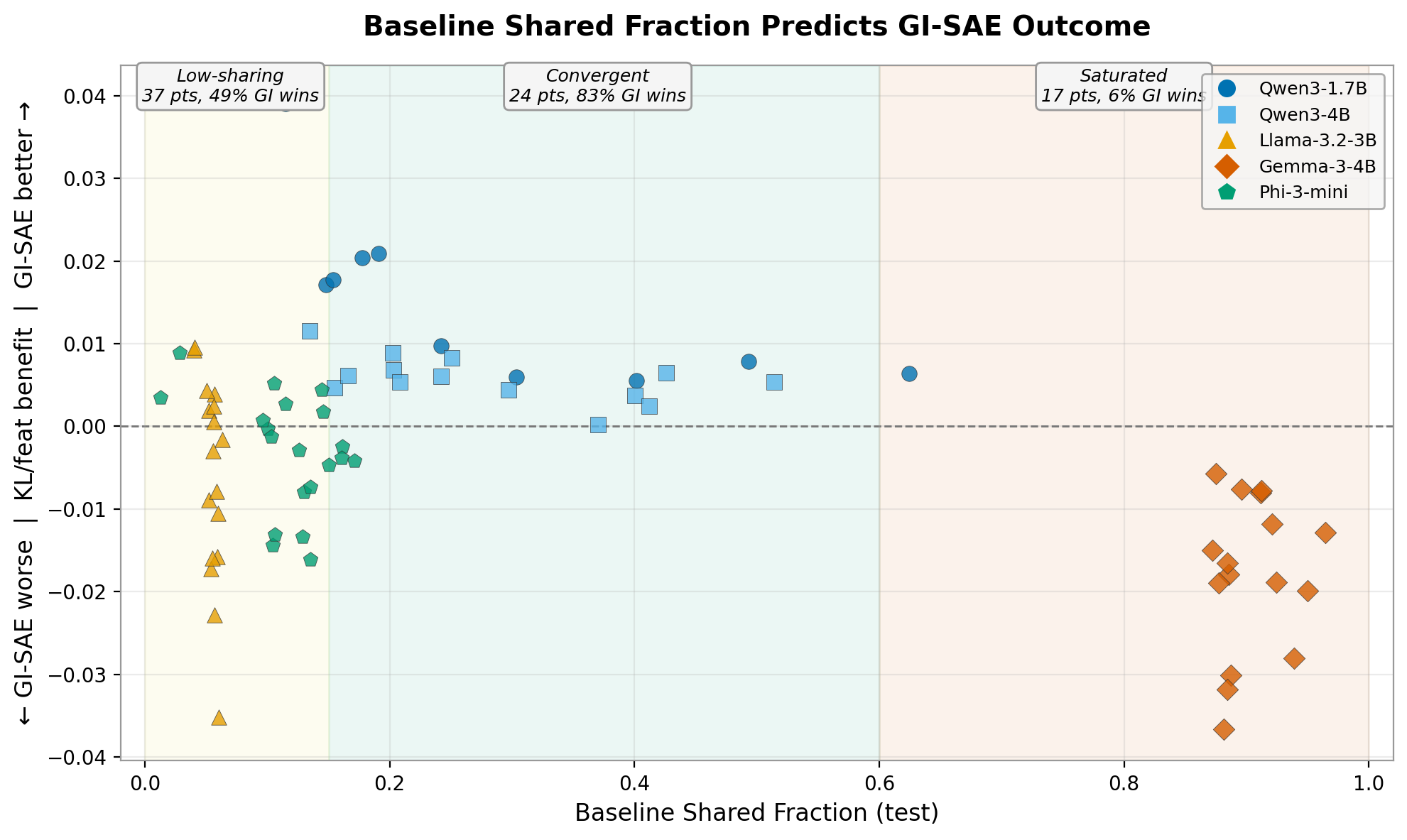}
  \caption{Phase diagram: GI-SAE KL/feature benefit vs.\ baseline
    shared fraction (test split).
    Points above the dashed line indicate GI-SAE wins.
    Three profiles emerge with distinct win rates.}
  \label{fig:profile-map}
\end{figure*}

When all 78 observations are stratified by baseline shared fraction
regardless of model identity (Figure~\ref{fig:profile-map}), a useful
empirical profile structure emerges:

\begin{table}[h!]
  \caption{GI-SAE win rates by baseline sharing profile (test split).
    95\% Wilson confidence intervals in brackets.}
  \label{tab:profile}
  \centering
  \footnotesize
  \setlength{\tabcolsep}{4pt}
  \begin{tabular}{lrrrr}
    \toprule
    Profile & Baseline shared & Obs. & Win rate & 95\% CI \\
    \midrule
    Low-sharing & $<$15\% & 37 & 49\% & [33, 64]\% \\
    Convergent  & 15--60\% & 24 & 83\% & [64, 93]\% \\
    Saturated   & $>$60\% & 17 & 6\% & [1, 27]\% \\
    \bottomrule
  \end{tabular}
\end{table}

GI-SAE reliably helps in the \emph{convergent} profile (enough baseline
sharing to build on), is counterproductive in the \emph{saturated}
profile (Gemma; swapping heavily-used shared features increases
disruption), and shows no systematic benefit in the \emph{low-sharing}
profile. Baseline shared fraction thus acts as a useful diagnostic in
this sample.

\subsection{Family Profiles of Cross-Language Feature Sharing}
\label{sec:results-family-profiles}

\begin{figure*}[t]
  \centering
  \includegraphics[width=\textwidth]{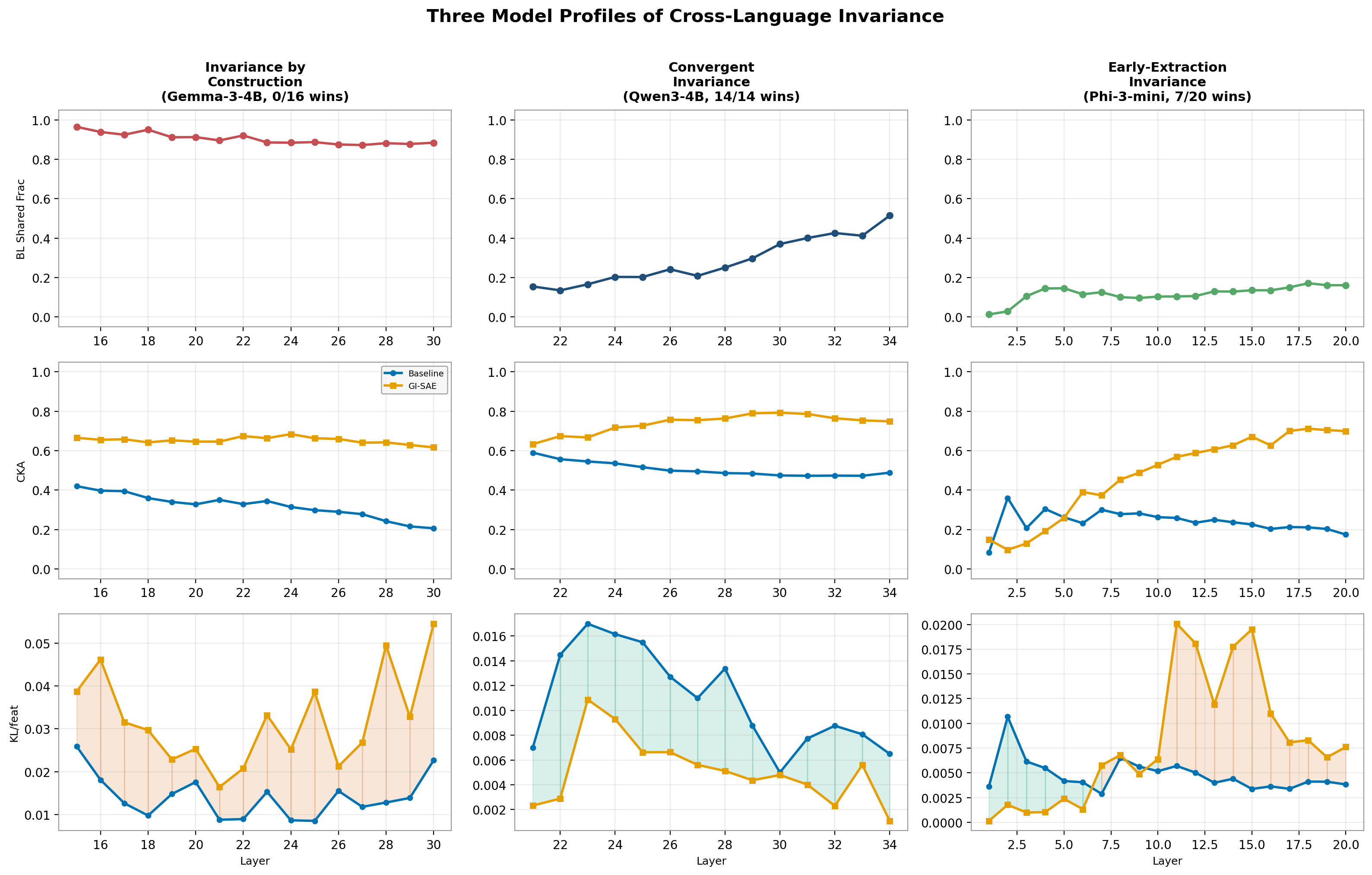}
  \caption{Three representative family profiles.
    \textbf{Top}: baseline shared fraction vs.\ layer.
    \textbf{Middle}: CKA (baseline blue, GI-SAE orange).
    \textbf{Bottom}: KL/feature (lower = greater functional
    interchangeability; shading marks GI-SAE advantage in green,
    disadvantage in orange).
    Left: Gemma (saturated). Center: Qwen3-4B (convergent).
    Right: Phi-3-mini (early-extraction).}
  \label{fig:taxonomy}
\end{figure*}

The five models reveal three recurring multilingual processing profiles
(Figure~\ref{fig:taxonomy}; Llama and Qwen3-1.7B patterns are shown in
the appendix).

\textbf{Qwen} shows progressive sharing: baseline shared fraction rises
monotonically with depth, and GI-SAE wins on all layers for both
models, which closely track one another.
\textbf{Gemma} already shares ${\sim}$90\% of features at baseline;
GI-SAE increases KL/feature at all 16 layers, supporting the
interpretation that the baseline already captures available
cross-language structure.
\textbf{Llama} achieves the largest CKA gains (up to $+0.67$) but
functional gains appear only in the deepest layers (L23--L27, with two
isolated early wins at L12--L13); mid-depth layers (L14--L22) show no
improvement.
\textbf{Phi} peaks in CKA at L9 (${\sim}$31\% depth); GI-SAE wins in
early layers (L1--L6) and loses from L7 onward, mirroring Llama in
reverse.

A qualitative audit of the most consistently shared features
(Appendix~\ref{app:audit}) shows that they encode interpretable
cross-language discourse structure: answer-boundary markers that fire on
sentence-final punctuation across 4--6 languages, answer-transition
phrases that fire on the verb introducing the answer regardless of
surface form, and reasoning-step markers active even at early layers
(L3 in Phi). These categories appear across the observed model profiles, though
the strongest shared features are concentrated near the
reasoning-to-answer transition rather than in mid-reasoning computation.

\subsection{Robustness and Training Health}
\label{sec:results-robustness}

Val and test KL/feature benefits are strongly aligned across the 78
layer observations (descriptive Pearson $r = 0.894$;
Appendix~\ref{app:ci}). A normalized-space patching sensitivity check
preserves the Qwen/convergent conclusion while showing that low/mixed
layers are more convention-sensitive
(Appendix~\ref{app:norm-sensitivity}).
SAE training diagnostics are in Appendix~\ref{app:training}: baseline
SAEs maintain $>$99\% alive features; GI-SAE remains mostly high but
drops to 84--88\% for Llama and late Qwen-1.7B layers.
Table~\ref{tab:summary} summarizes the five-model sweep.

\begin{table*}[t]
  \caption{Five-model summary (test split).
    95\% Wilson CIs for win rates. Full per-layer results in
    Appendix~\ref{app:layer-results}.}
  \label{tab:summary}
  \centering
  \footnotesize
  \begin{tabular}{lrrrrrrr}
    \toprule
    Model & Layers & Test wins & 95\% CI & CKA $\Delta$ & BL shared & GI shared & Profile \\
    \midrule
    Qwen3-1.7B & 11 & 11/11 & [74, 100]\% & +0.19 & 11--62\% & 16--67\% & Conv \\
    Qwen3-4B   & 14 & 14/14 & [78, 100]\% & +0.23 & 13--51\% & 24--69\% & Conv \\
    Llama-3.2-3B & 17 & 7/17 & [22, 64]\% & +0.57 & 4--6\% & 1--17\% & Mixed \\
    Gemma-3-4B & 16 & 0/16 & [0, 19]\% & +0.33 & 87--96\% & 85--94\% & Sat \\
    Phi-3-mini & 20 & 7/20 & [18, 57]\% & +0.24 & 1--17\% & 9--39\% & Mixed \\
    \bottomrule
  \end{tabular}
\end{table*}
% ---- end parts/main/05_results ----
% ---- inlined from parts/main/06_discussion ----
\section{Discussion}
\label{sec:discussion}

\paragraph{Interpretive scope of GI-SAE}
GI-SAE is informative even when it does not improve functional
interchangeability.
In the \emph{saturated} profile (Gemma), GI-SAE's inability to improve
sharing supports the interpretation that the baseline SAE already captures the
cross-language structure present in the model; the contrastive loss has
nothing to add.
In the \emph{low-sharing} profile (Llama early layers), GI-SAE achieves
large geometric similarity gains (CKA~$\Delta$ up to +0.67) without
improving functional interchangeability, showing that CKA and Jaccard
measure representation similarity, not whether shared features can be
swapped without disrupting output.

This gap is a general methodological insight: wherever CKA is used to
infer functional equivalence, the same gap may apply.
Baseline shared fraction (Table~\ref{tab:profile}) stratifies whether
contrastive training helps in this sample.

\subsection{Limitations and Future Directions}
\label{sec:discussion-limitations}

\paragraph{Limitations.}
The taxonomy is descriptive: each non-Qwen family is one checkpoint;
Qwen provides only a within-generation scale check.
(i)~MGSM only; may not generalize to harder math, code, or
non-mathematical domains.
(ii)~Llama and Phi have ${\sim}$60 valid problems vs.\ 151--202 for
Qwen, limiting statistical power.
(iii)~Models up to 4B; scaling to 8B+ reduces the active dictionary
fraction.
(iv)~Fixed hyperparameters ($K{=}128$, $4{\times}$, $\tau{=}0.1$,
$w{=}1.0$) across all models; $w{=}1.0$ was selected from a single
Qwen3-1.7B L20 comparison and held fixed as a common intervention
strength (Appendix~\ref{app:training}).
(v)~Contrastive positives are problem-level, not offset-matched;
this may align answer-level rather than step-level information.
(vi)~One sampled trace per problem-language pair; multiple traces
would separate representation effects from sampling variance.

\paragraph{Future directions.}
(i)~Within-family replications (e.g., Gemma-2-2B, Llama-3.2-1B) to
test whether sharing profiles are family-stable.
(ii)~CKA-only lightweight surveys across additional models to extend
the phase diagram without full pipeline cost.
(iii)~Extension beyond mathematical reasoning to code, logical
inference, and general instruction following.
(iv)~Adaptive contrastive weight or temperature scheduling conditioned
on baseline shared fraction.
(v)~Investigation of what architectural or training choices produce
high default feature sharing (Gemma) vs.\ progressive building (Qwen)
vs.\ persistent language-specificity (Llama).
% ---- end parts/main/06_discussion ----
% ---- inlined from parts/main/07_conclusion ----
\section{Conclusion}
\label{sec:conclusion}

We introduced GI-SAE, a contrastive sparse autoencoder, and applied it
alongside a baseline SAE to five models from four families under a
common 6-language mathematical reasoning protocol. Cross-language
feature sharing varies strongly across models (progressive in
Qwen, saturated in Gemma, mixed in Llama and Phi) and geometric
similarity does not reliably predict functional interchangeability:
GI-SAE improves CKA at nearly every layer, yet KL/feature improves
only in the convergent profile (83\% win rate) and not in the saturated
profile (6\%).

These findings show that geometric similarity should not be used as a
proxy for functional interchangeability, and that cross-language
feature sharing must be analyzed on a per-model and per-architecture basis.
% ---- end parts/main/07_conclusion ----
% ---- end parts/main/main ----

% Acknowledgments must remain omitted for double-blind review.
% \section*{Acknowledgments}
% \input{parts/main/08_acknowledgments}

\section*{Impact Statement}
This work is a foundational mechanistic-interpretability study of internal
representations in existing open-source language models. It does not introduce
new model capabilities, generate content, or release new pretrained models or
datasets with direct deployment implications. Its potential positive impact is
improved understanding of multilingual model behavior, which may support future
interpretability and auditing research. The study uses public data and involves
no human subjects or private information. We do not foresee specific negative
societal impacts beyond those generally associated with research on existing
large language models.

\bibliography{bib}
\bibliographystyle{icml2026}

\newpage
\appendix
\onecolumn
% ---- inlined from parts/appendix/appendix ----
\section{Full Per-Model Layer Profiles}
\label{app:profiles}

Figure~\ref{fig:taxonomy} in the main text shows depth profiles for
Gemma, Qwen3-4B, and Phi-3-mini. Here we present the remaining two
models: Llama-3.2-3B and Qwen3-1.7B.

\paragraph{Llama-3.2-3B (low-sharing profile).}
Baseline shared fraction remains at 4--6\% across all 17 swept layers
(L11--L27). GI-SAE increases sharing to 1--17\%, with gradual growth
toward deeper layers. CKA improvements are the largest in the study
(up to $+0.67$), yet sustained KL/feature improvements appear only in
the deepest layers (L23--L27), with two isolated early wins (L12--L13).
Mid-depth layers (L14--L22) show geometric gain without functional
benefit, indicating that the model processes languages through largely
separate pathways at those depths.

\paragraph{Qwen3-1.7B (convergent profile).}
Shared fraction increases monotonically from 11\% at L16 to 62\% at L26,
closely tracking Qwen3-4B and confirming within-family consistency under
scale change. GI-SAE wins on all 11 test layers (100\% win rate,
95\% CI [74\%, 100\%]). Peak shared fraction reaches 67\% under GI-SAE.
The 1.7B model achieves slightly lower peak CKA (0.69 vs.\ 0.79 for
4B) but follows the same progressive-sharing trajectory.

%% When figures are generated, uncomment:
% \begin{figure}[h]
%   \centering
%   \includegraphics[width=\linewidth]{colorblind/fig3_llama_profile.png}
%   \caption{Llama-3.2-3B layer profile. Top: baseline shared fraction.
%     Middle: CKA (baseline blue, GI-SAE orange). Bottom: KL/feature.}
%   \label{fig:app-llama-profile}
% \end{figure}
%
% \begin{figure}[h]
%   \centering
%   \includegraphics[width=\linewidth]{colorblind/fig3_qwen17b_profile.png}
%   \caption{Qwen3-1.7B layer profile. Same format as above.}
%   \label{fig:app-qwen17b-profile}
% \end{figure}

\section{Full Layer-by-Layer Results}
\label{app:layer-results}

Tables~\ref{tab:app-qwen17b}--\ref{tab:app-phi} report all 78
(model, layer) observations. Shared fraction is $|S|/K$ where
$K{=}128$. KL/feat ($\times 10^3$) is the test-split normalized metric
(Equation~\ref{eq:kl-per-feat}). ``Win'' (\checkmark) indicates
$\text{KL/feat}_\text{GI} < \text{KL/feat}_\text{BL}$.

\begin{table}[h]
  \caption{Qwen3-1.7B layer-by-layer results (test split, L16--L26).
    Shared\% = $|S|/128$. KL/feat $\times 10^3$.
    GI-SAE wins 11/11 layers.}
  \label{tab:app-qwen17b}
  \centering
  \scriptsize
  \begin{tabular}{lrrrrrrl}
    \toprule
    Layer & BL Sh.\% & GI Sh.\% & BL KL/f & GI KL/f & BL AFR & GI AFR & Win \\
    \midrule
    L16 & 11.5 & 17.4 & 42.0 & 2.9 & 23.6 & 6.7 & \checkmark \\
    L17 & 14.8 & 36.5 & 46.3 & 29.2 & 25.8 & 30.0 & \checkmark \\
    L18 & 15.4 & 40.0 & 37.4 & 19.6 & 25.8 & 24.4 & \checkmark \\
    L19 & 13.4 & 16.3 & 43.2 & 3.3 & 24.7 & 7.8 & \checkmark \\
    L20 & 17.8 & 45.3 & 27.0 & 6.6 & 26.9 & 16.4 & \checkmark \\
    L21 & 19.1 & 43.6 & 26.5 & 5.6 & 23.3 & 18.1 & \checkmark \\
    L22 & 24.2 & 42.7 & 13.9 & 4.2 & 21.1 & 17.2 & \checkmark \\
    L23 & 30.3 & 51.3 & 9.5 & 3.6 & 17.2 & 16.1 & \checkmark \\
    L24 & 40.2 & 59.4 & 8.1 & 2.6 & 27.8 & 15.6 & \checkmark \\
    L25 & 49.3 & 64.3 & 10.3 & 2.5 & 31.4 & 15.6 & \checkmark \\
    L26 & 62.4 & 66.7 & 8.5 & 2.1 & 36.4 & 21.9 & \checkmark \\
    \bottomrule
  \end{tabular}
\end{table}

\begin{table}[h]
  \caption{Qwen3-4B layer-by-layer results (test split, L21--L34).
    GI-SAE wins 14/14 layers.}
  \label{tab:app-qwen4b}
  \centering
  \scriptsize
  \begin{tabular}{lrrrrrrl}
    \toprule
    Layer & BL Sh.\% & GI Sh.\% & BL KL/f & GI KL/f & BL AFR & GI AFR & Win \\
    \midrule
    L21 & 15.5 & 26.6 & 7.0 & 2.3 & 14.2 & 9.0 & \checkmark \\
    L22 & 13.5 & 24.0 & 14.5 & 2.9 & 22.2 & 10.1 & \checkmark \\
    L23 & 16.6 & 40.7 & 17.0 & 10.9 & 23.7 & 23.7 & \checkmark \\
    L24 & 20.3 & 42.6 & 16.1 & 9.3 & 26.5 & 21.3 & \checkmark \\
    L25 & 20.3 & 42.9 & 15.5 & 6.6 & 25.4 & 20.0 & \checkmark \\
    L26 & 24.2 & 44.7 & 12.7 & 6.6 & 23.7 & 20.2 & \checkmark \\
    L27 & 20.8 & 44.2 & 11.0 & 5.6 & 20.6 & 26.5 & \checkmark \\
    L28 & 25.1 & 48.7 & 13.4 & 5.1 & 27.7 & 21.7 & \checkmark \\
    L29 & 29.7 & 55.7 & 8.8 & 4.4 & 23.0 & 14.4 & \checkmark \\
    L30 & 37.0 & 55.5 & 5.0 & 4.8 & 13.3 & 16.3 & \checkmark \\
    L31 & 40.0 & 61.8 & 7.7 & 4.0 & 28.4 & 18.1 & \checkmark \\
    L32 & 42.6 & 62.0 & 8.8 & 2.3 & 37.8 & 16.3 & \checkmark \\
    L33 & 41.2 & 58.1 & 8.1 & 5.6 & 35.3 & 25.4 & \checkmark \\
    L34 & 51.4 & 68.7 & 6.5 & 1.1 & 34.6 & 20.0 & \checkmark \\
    \bottomrule
  \end{tabular}
\end{table}

\begin{table}[h]
  \caption{Llama-3.2-3B layer-by-layer results (test split, L11--L27).
    GI-SAE wins 7/17 layers: L12--L13 and L23--L27.}
  \label{tab:app-llama}
  \centering
  \scriptsize
  \begin{tabular}{lrrrrrrl}
    \toprule
    Layer & BL Sh.\% & GI Sh.\% & BL KL/f & GI KL/f & BL AFR & GI AFR & Win \\
    \midrule
    L11 & 5.7 & 1.1 & 3.9 & 26.7 & 8.9 & 5.9 & --- \\
    L12 & 5.2 & 4.8 & 5.3 & 3.3 & 8.9 & 5.2 & \checkmark \\
    L13 & 5.7 & 6.8 & 10.4 & 6.5 & 10.4 & 7.4 & \checkmark \\
    L14 & 5.9 & 8.7 & 10.1 & 25.9 & 16.3 & 14.1 & --- \\
    L15 & 6.0 & 9.7 & 12.3 & 22.9 & 12.6 & 17.0 & --- \\
    L16 & 5.2 & 10.0 & 18.2 & 27.1 & 16.3 & 20.0 & --- \\
    L17 & 6.1 & 9.2 & 14.4 & 49.5 & 11.1 & 25.9 & --- \\
    L18 & 5.4 & 9.4 & 16.5 & 33.7 & 14.8 & 23.0 & --- \\
    L19 & 5.5 & 10.5 & 15.7 & 31.7 & 13.3 & 26.7 & --- \\
    L20 & 5.8 & 11.4 & 15.6 & 23.5 & 15.6 & 22.2 & --- \\
    L21 & 6.3 & 11.7 & 15.0 & 16.6 & 17.8 & 22.2 & --- \\
    L22 & 5.6 & 11.9 & 12.8 & 15.8 & 13.3 & 20.0 & --- \\
    L23 & 5.6 & 13.4 & 12.7 & 12.1 & 14.1 & 18.5 & \checkmark \\
    L24 & 5.6 & 13.4 & 12.4 & 9.9 & 15.6 & 13.3 & \checkmark \\
    L25 & 5.1 & 14.9 & 9.4 & 5.1 & 14.1 & 17.0 & \checkmark \\
    L26 & 4.0 & 15.7 & 12.4 & 3.2 & 15.6 & 10.4 & \checkmark \\
    L27 & 4.1 & 17.2 & 12.6 & 3.0 & 14.1 & 8.9 & \checkmark \\
    \bottomrule
  \end{tabular}
\end{table}

\begin{table}[h]
  \caption{Gemma-3-4B layer-by-layer results (test split, L15--L30).
    GI-SAE wins 0/16 layers.}
  \label{tab:app-gemma}
  \centering
  \scriptsize
  \begin{tabular}{lrrrrrrl}
    \toprule
    Layer & BL Sh.\% & GI Sh.\% & BL KL/f & GI KL/f & BL AFR & GI AFR & Win \\
    \midrule
    L15 & 96.5 & 94.1 & 25.9 & 38.7 & 63.3 & 76.7 & --- \\
    L16 & 93.9 & 94.4 & 18.1 & 46.2 & 61.5 & 77.0 & --- \\
    L17 & 92.5 & 93.2 & 12.6 & 31.5 & 42.4 & 53.0 & --- \\
    L18 & 95.0 & 92.8 & 9.8 & 29.7 & 30.6 & 55.2 & --- \\
    L19 & 91.2 & 93.5 & 14.8 & 22.9 & 30.0 & 56.4 & --- \\
    L20 & 91.2 & 94.1 & 17.6 & 25.3 & 55.8 & 53.9 & --- \\
    L21 & 89.6 & 91.6 & 8.8 & 16.4 & 27.0 & 28.2 & --- \\
    L22 & 92.1 & 92.8 & 8.9 & 20.8 & 26.4 & 42.4 & --- \\
    L23 & 88.5 & 92.3 & 15.3 & 33.2 & 26.1 & 42.7 & --- \\
    L24 & 88.4 & 86.1 & 8.6 & 25.2 & 26.7 & 42.7 & --- \\
    L25 & 88.7 & 86.7 & 8.5 & 38.7 & 17.6 & 52.4 & --- \\
    L26 & 87.5 & 85.1 & 15.5 & 21.3 & 21.5 & 38.8 & --- \\
    L27 & 87.2 & 89.1 & 11.8 & 26.8 & 19.7 & 33.6 & --- \\
    L28 & 88.2 & 89.9 & 12.8 & 49.5 & 21.8 & 44.2 & --- \\
    L29 & 87.8 & 89.3 & 13.9 & 32.8 & 20.0 & 39.4 & --- \\
    L30 & 88.4 & 87.1 & 22.7 & 54.5 & 25.5 & 45.2 & --- \\
    \bottomrule
  \end{tabular}
\end{table}

\begin{table}[h]
  \caption{Phi-3-mini layer-by-layer results (test split, L1--L20).
    GI-SAE wins 7/20 layers, all in L1--L6 and L9.}
  \label{tab:app-phi}
  \centering
  \scriptsize
  \begin{tabular}{lrrrrrrl}
    \toprule
    Layer & BL Sh.\% & GI Sh.\% & BL KL/f & GI KL/f & BL AFR & GI AFR & Win \\
    \midrule
    L1 & 1.3 & 17.6 & 3.6 & 0.1 & 4.7 & 3.3 & \checkmark \\
    L2 & 2.8 & 9.2 & 10.7 & 1.8 & 10.0 & 10.0 & \checkmark \\
    L3 & 10.6 & 10.0 & 6.1 & 1.0 & 12.0 & 4.7 & \checkmark \\
    L4 & 14.5 & 11.6 & 5.5 & 1.0 & 9.3 & 6.7 & \checkmark \\
    L5 & 14.6 & 10.8 & 4.2 & 2.4 & 10.7 & 6.7 & \checkmark \\
    L6 & 11.5 & 11.2 & 4.0 & 1.3 & 11.3 & 4.7 & \checkmark \\
    L7 & 12.6 & 23.7 & 2.9 & 5.8 & 10.0 & 12.7 & --- \\
    L8 & 10.0 & 28.7 & 6.5 & 6.8 & 12.0 & 10.7 & --- \\
    L9 & 9.6 & 28.3 & 5.6 & 4.9 & 10.7 & 9.3 & \checkmark \\
    L10 & 10.3 & 27.1 & 5.2 & 6.4 & 8.0 & 9.3 & --- \\
    L11 & 10.4 & 24.9 & 5.7 & 20.1 & 8.0 & 9.3 & --- \\
    L12 & 10.6 & 25.0 & 5.0 & 18.1 & 7.3 & 12.0 & --- \\
    L13 & 13.0 & 31.2 & 4.0 & 11.9 & 7.3 & 18.0 & --- \\
    L14 & 12.9 & 27.2 & 4.4 & 17.8 & 9.3 & 16.7 & --- \\
    L15 & 13.5 & 28.4 & 3.4 & 19.5 & 6.7 & 18.0 & --- \\
    L16 & 13.5 & 29.5 & 3.6 & 11.0 & 10.0 & 19.3 & --- \\
    L17 & 15.0 & 31.3 & 3.4 & 8.1 & 12.0 & 14.0 & --- \\
    L18 & 17.1 & 33.3 & 4.1 & 8.3 & 10.7 & 12.7 & --- \\
    L19 & 16.1 & 34.5 & 4.1 & 6.6 & 10.7 & 14.7 & --- \\
    L20 & 16.1 & 39.0 & 3.8 & 7.6 & 13.3 & 12.0 & --- \\
    \bottomrule
  \end{tabular}
\end{table}

\section{Causal Patching Controls}
\label{app:controls}

To test whether low cross-language KL divergence reflects genuine
functional interchangeability, we add two control conditions to the
standard patching protocol, using the same trained SAEs (no retraining):
\textbf{different-problem donor} (features from a different math
problem, same source language) and \textbf{random-value donor}
(same shared feature indices, but activation values permuted across
those indices). If same-problem cross-language patches are less
disruptive than both controls, the shared features encode
problem-specific reasoning, not generic structure.

Table~\ref{tab:controls} reports KL/shared feature for 7
representative layers across all 5 models and both SAE variants.

\begin{table}[h]
  \caption{Causal patching controls (test split). SP = same-problem
    cross-language, DP = different-problem, RV = random-value.
    Pass ($\checkmark$) = control KL/feat $>$ SP KL/feat.}
  \label{tab:controls}
  \centering
  \scriptsize
  \begin{tabular}{llllrrrcc}
    \toprule
    Model & Layer & Profile & Variant & SP & DP & RV & DP$>$SP & RV$>$SP \\
    \midrule
    Qwen-4B & L29 & Conv & BL & 0.0088 & 0.0114 & 0.0127 & \checkmark & \checkmark \\
    & & & GI & 0.0044 & 0.0071 & 0.0111 & \checkmark & \checkmark \\
    Qwen-1.7B & L24 & Conv & BL & 0.0081 & 0.0092 & 0.0194 & \checkmark & \checkmark \\
    & & & GI & 0.0026 & 0.0032 & 0.0064 & \checkmark & \checkmark \\
    Llama & L17 & Low & BL & 0.0144 & 0.0206 & 0.0164 & \checkmark & \checkmark \\
    & & & GI & 0.0495 & 0.0741 & 0.0519 & \checkmark & \checkmark \\
    Llama & L25 & Low & BL & 0.0094 & 0.0106 & 0.0125 & \checkmark & \checkmark \\
    & & & GI & 0.0051 & 0.0080 & 0.0056 & \checkmark & \checkmark \\
    Gemma & L21 & Sat & BL & 0.0088 & 0.0081 & 0.0645 & --- & \checkmark \\
    & & & GI & 0.0160 & 0.0191 & 0.0984 & \checkmark & \checkmark \\
    Phi & L3 & Low & BL & 0.0061 & 0.0019 & 0.0063 & --- & \checkmark \\
    & & & GI & 0.0010 & 0.0011 & 0.0011 & \checkmark & \checkmark \\
    Phi & L15 & Low & BL & 0.0034 & 0.0042 & 0.0041 & \checkmark & \checkmark \\
    & & & GI & 0.0195 & 0.0333 & 0.0192 & \checkmark & --- \\
    \midrule
    \multicolumn{4}{l}{Pass rate (BL)} & & 5/7 & 7/7 & 71\% & 100\% \\
    \multicolumn{4}{l}{Pass rate (GI)} & & 7/7 & 6/7 & 100\% & 86\% \\
    \multicolumn{4}{l}{\textbf{Overall}} & & \textbf{12/14} & \textbf{13/14} & \textbf{86\%} & \textbf{93\%} \\
    \bottomrule
  \end{tabular}
\end{table}

Overall pass rate is 89\% (25/28). The three failures are explainable
edge cases: (1)~Gemma BL different-problem (0.0081 vs.\ 0.0088):
saturated profile shares ${\sim}$90\% of features regardless of problem,
so different-problem donors are nearly indistinguishable; (2)~Phi L3
BL different-problem (0.0019 vs.\ 0.0061): very low baseline sharing
(${\sim}$13 features) yields too few patched features for
problem-specificity to manifest; (3)~Phi L15 GI random-value
(0.0192 vs.\ 0.0195): statistical tie (ratio 0.98$\times$).

GI-SAE features show stronger problem-specificity than baseline
features: GI passes the different-problem control 100\% (7/7) vs.\
baseline 71\% (5/7).

\section{Pipeline and Implementation Details}
\label{app:patching-details}

\subsection{Per-Model Experimental Pipeline}

For each model, the pipeline proceeds through six stages:
(1)~activation extraction: replay all valid reasoning traces and record
residual-stream vectors at each target layer for every token;
(2)~SAE training: train baseline SAE and GI-SAE on train-split
activations at each layer;
(3)~geometric evaluation: compute CKA and Jaccard similarity between
language pairs on validation activations encoded through each SAE;
(4)~causal patching: run Algorithm~1 for all (problem, language pair)
combinations on both validation and test splits;
(5)~aggregation: collect per-layer results into a single results file;
(6)~analysis: generate per-model comparison plots.
Each model's pipeline takes approximately 7--10 hours; the complete
five-model sweep requires ${\sim}$60 GPU-hours on a single NVIDIA
RTX 5090.

\subsection{Causal Patching Protocol}

For each valid reasoning trace, we replay the exact generated token
sequence (replaying exact generated tokens) using
TransformerLens~\citep{nanda2022transformerlens}. At the target layer
and a backward-aligned token position, we record the target
residual-stream vector $r_\text{tgt}$ and the donor residual-stream
vector $r_\text{src}$ (from a different language's trace of the same
problem).

Both vectors are encoded through the trained SAE \emph{without}
z-score normalization, because the patched vector is injected back into
the model's forward pass at the original activation scale. Features
active in both encodings ($f_{\text{tgt},j} > 0$ and
$f_{\text{src},j} > 0$) constitute the shared set $S$.

The patched residual is constructed as:
\[
r_\text{patched} = \operatorname{decode}(f_\text{patched})
  + (r_\text{tgt} - \operatorname{decode}(f_\text{tgt})),
\]
where $f_\text{patched}$ equals $f_\text{tgt}$ except at indices in $S$,
which take donor values. Adding the reconstruction error
$(r_\text{tgt} - \operatorname{decode}(f_\text{tgt}))$ ensures the
intervention modifies only the SAE feature subspace and preserves the
residual component.

The patched vector replaces $r_\text{tgt}$ at the same layer and
position via a TransformerLens hook, and the forward pass continues
over all remaining layers. KL divergence is measured between the clean
and patched next-token distributions at the \emph{final reasoning-token
position} (backward offset $-1$), after full downstream propagation.
Autoregressive flip rate (AFR) records whether the top-1 next-token
prediction at that position changes. AFR is a local
prediction-sensitivity metric, not a regenerated final-answer
comparison.

When $|S| = 0$ (no shared active features), the intervention is skipped
for that (problem, language pair, position) triple. Same-language pairs
serve as a near-zero disruption control: because both encodings are
identical (same trace replayed), patching produces no change.

Backward offsets are processed independently; results at each offset are
averaged across all valid (problem, language pair) combinations. The
primary metrics reported in the main text aggregate across all offsets.

\section{Dataset Filtering and Prompts}
\label{app:prompts}

\subsection{Prompts}

Each problem is presented via a system prompt instructing the model to
reason in the target language and a user prompt containing the problem.
Table~\ref{tab:prompts} shows the system prompts for all six
languages.\footnote{Russian and Chinese prompts are shown in
romanized transliteration due to typesetting constraints.}

\begin{table}[h]
  \caption{System prompts by language.}
  \label{tab:prompts}
  \centering
  \small
  \begin{tabular}{lp{10.5cm}}
    \toprule
    Lang & System prompt \\
    \midrule
    en & You are a math problem solver. Reason step by step in English.
         At the end, write your answer as JSON: \{``answer'': N\}. \\
    de & Du bist ein mathematischer Probleml\"oser. Denke Schritt f\"ur
         Schritt auf Deutsch. Schreibe am Ende deine Antwort als JSON:
         \{``Antwort'': N\}. \\
    fr & Tu es un r\'esolveur de probl\`emes math\'ematiques. Raisonne
         \'etape par \'etape en fran\c{c}ais. \`A la fin, \'ecris ta
         r\'eponse en JSON~: \{``r\'eponse'': N\}. \\
    es & Eres un resolutor de problemas matem\'aticos. Razona paso a
         paso en espa\~nol. Al final, escribe tu respuesta como JSON:
         \{``respuesta'': N\}. \\
    ru & [Romanized] Ty reshatel' matematicheskikh zadach.
         Rassuzhdaj poshagovo na russkom yazyke. V kontse zapishi
         otvet v formate JSON: \{``otvet'': N\}. \\
    zh & [Romanized] Ni shi yige shuxue wenti qiujieqi.
         Yong zhongwen zhubi tuili. Zuihou jiang da'an xiecheng
         JSON geshi: \{``da'an'': N\}. \\
    \bottomrule
  \end{tabular}
\end{table}

User prompts follow the template: ``Solve the problem. Think step by
step: \{problem\}. Provide your final numerical answer strictly as JSON
without units: \{\textit{key}: VALUE\}.'' Equivalent translations are
used for each language with localized JSON answer keys.

\subsection{Reasoning Trace Extraction}

Two strategies are used depending on model architecture:
\begin{enumerate}
  \item \textbf{Qwen models}: explicit reasoning inside
    \texttt{<think>}$\ldots$\texttt{</think>} tags is extracted via
    regex.
  \item \textbf{Gemma, Llama, Phi}: reasoning is defined as all text
    preceding the final JSON answer block
    (e.g., \texttt{\{``answer'': 42\}}). Trailing markdown code fences
    are stripped.
\end{enumerate}
In both cases, the extracted \texttt{think\_text} contains only
reasoning with no JSON answer block or markup, though natural-language
answer statements may remain in the reasoning span.

\subsection{Filtering and Validity}

A problem is retained only if the model solves it correctly \emph{and}
produces an extractable reasoning trace in all six languages. All
results in this paper are therefore conditional on successful
multilingual reasoning. This design isolates the setting where
cross-language feature comparison is meaningful, but should not be
interpreted as estimating feature sharing over the full MGSM
distribution.

Table~\ref{tab:models} in the main text reports valid problem counts
per model: 202 (Qwen-4B), 151 (Qwen-1.7B), 137 (Gemma), 63 (Phi),
60 (Llama). The large variance reflects differences in multilingual
mathematical competence.

\subsection{Inference Parameters}

\begin{table}[h]
  \caption{Generation hyperparameters (all models).}
  \label{tab:inference}
  \centering
  \begin{tabular}{lr}
    \toprule
    Parameter & Value \\
    \midrule
    Max new tokens & 2048 \\
    Temperature & 0.6 \\
    Top-$p$ & 0.95 \\
    Top-$k$ & 20 \\
    Sampling & Yes \\
    Seed & $42 + \text{problem index}$ \\
    Padding & Left \\
    Precision & bfloat16 \\
    \bottomrule
  \end{tabular}
\end{table}

\section{SAE Training Diagnostics}
\label{app:training}

Table~\ref{tab:training-diag} summarizes training health across all
models. Baseline SAE achieves $>$99\% alive features everywhere.
GI-SAE alive fraction remains above 90\% for most layers but drops to
84--88\% for Llama and late-layer Qwen-1.7B, where the contrastive and
reconstruction objectives are in strongest tension. GI-SAE validation
loss is typically 1.5--2.0$\times$ baseline, with spikes above
3$\times$ at select Qwen layers (e.g., L16 and L19 for Qwen-1.7B).
These spikes do not affect downstream evaluation: affected layers
still produce SAEs with $>$90\% active features.

\begin{table}[h]
  \caption{SAE training diagnostics (representative layers).
    Val loss is reconstruction MSE on the validation split.
    Alive\% is the fraction of $d_\text{sae}$ features firing
    on at least one validation sample.}
  \label{tab:training-diag}
  \centering
  \scriptsize
  \begin{tabular}{llrrrrl}
    \toprule
    Model & Layer & BL Val & GI Val & Ratio & GI Alive\% & Note \\
    \midrule
    Qwen-1.7B & L16 & 0.213 & 0.673 & 3.2$\times$ & 99.7 & High ratio \\
    & L19 & 0.153 & 0.649 & 4.2$\times$ & 99.6 & High ratio \\
    & L22 & 0.161 & 0.313 & 1.9$\times$ & 89.4 & \\
    & L26 & 0.193 & 0.343 & 1.8$\times$ & 88.7 & \\
    \midrule
    Qwen-4B & L21 & 0.225 & 0.684 & 3.0$\times$ & 98.7 & High ratio \\
    & L22 & 0.197 & 0.675 & 3.4$\times$ & 98.6 & High ratio \\
    & L29 & 0.167 & 0.314 & 1.9$\times$ & 94.3 & \\
    & L34 & 0.175 & 0.323 & 1.8$\times$ & 94.3 & \\
    \midrule
    Llama & L11 & 0.306 & 0.385 & 1.3$\times$ & 91.7 & \\
    & L17 & 0.253 & 0.356 & 1.4$\times$ & 86.4 & \\
    & L27 & 0.247 & 0.363 & 1.5$\times$ & 89.9 & \\
    \midrule
    Gemma & L15 & 0.259 & 0.357 & 1.4$\times$ & 96.4 & \\
    & L21 & 0.194 & 0.309 & 1.6$\times$ & 95.9 & \\
    & L30 & 0.202 & 0.364 & 1.8$\times$ & 95.9 & \\
    \midrule
    Phi & L1 & 0.128 & 0.251 & 2.0$\times$ & 99.0 & \\
    & L9 & 0.226 & 0.373 & 1.7$\times$ & 96.8 & \\
    & L20 & 0.210 & 0.309 & 1.5$\times$ & 96.1 & \\
    \bottomrule
  \end{tabular}
\end{table}

\paragraph{Contrastive weight selection.}
We use $w{=}1.0$ as a fixed contrastive weight rather than tuning $w$
per model or layer. This value was selected from a preliminary
comparison on Qwen3-1.7B L20 using the merged single-pass
implementation. At $w{=}1.0$, CKA improved from 0.503 to 0.673 and
KL/feature improved from 0.0070 to 0.0003 relative to the baseline,
while validation loss increased only from 31.60 to 32.29. A smaller
weight ($w{=}0.1$) preserved reconstruction loss but produced weaker
invariance (CKA 0.492, KL/feature 0.0055). We therefore fixed
$w{=}1.0$ across the five-model sweep; adaptive weighting remains
future work.

\section{KL Decomposition}
\label{app:kl-decomp}

Table~\ref{tab:kl-decomp} decomposes the KL/shared feature metric into
its numerator (raw cross-language KL) and denominator (mean shared
feature count $|S|$), aggregated by sharing profile. This verifies
that the metric is not an artifact of division by $|S|$.

\begin{table}[h]
  \caption{KL decomposition by sharing profile (test split, all layers).
    Raw KL = mean cross-language KL divergence.
    $|S|$ = mean shared feature count. $|S|/128$ = shared fraction.
    KL/feat = primary metric. AFR = autoregressive flip rate.}
  \label{tab:kl-decomp}
  \centering
  \begin{tabular}{llrrrrr}
    \toprule
    Profile & Variant & Raw KL & $|S|$ & $|S|/128$ & KL/feat & AFR \\
    \midrule
    Low-sharing & BL & 0.198 & 15.2 & 11.9\% & 0.0124 & 15.2\% \\
    & GI & 0.301 & 30.8 & 24.0\% & 0.0114 & 14.9\% \\
    Convergent & BL & 0.426 & 53.1 & 41.5\% & 0.0080 & 28.2\% \\
    & GI & \textbf{0.249} & \textbf{77.0} & \textbf{60.1\%} & \textbf{0.0033} & \textbf{17.9\%} \\
    Saturated & BL & 1.581 & 113.6 & 88.8\% & 0.0138 & 32.5\% \\
    & GI & 3.516 & 114.4 & 89.3\% & 0.0303 & 47.3\% \\
    \bottomrule
  \end{tabular}
\end{table}

In the \textbf{convergent} profile, GI-SAE wins because both
components improve: raw KL drops ($0.43 \to 0.25$, $-41\%$) and sharing
increases ($42\% \to 60\%$, $+43\%$). In the \textbf{low-sharing}
profile, GI-SAE doubles sharing ($12\% \to 24\%$) but also increases
raw KL ($0.20 \to 0.30$), yielding only marginal KL/feat improvement.
In the \textbf{saturated} profile, sharing is ceiling-ed ($89\% \approx
89\%$) while raw KL more than doubles ($1.58 \to 3.52$), making the
GI-SAE intervention destructive.

\section{Statistical Uncertainty}
\label{app:ci}

\begin{figure*}[h!]
  \centering
  \includegraphics[width=0.9\linewidth]{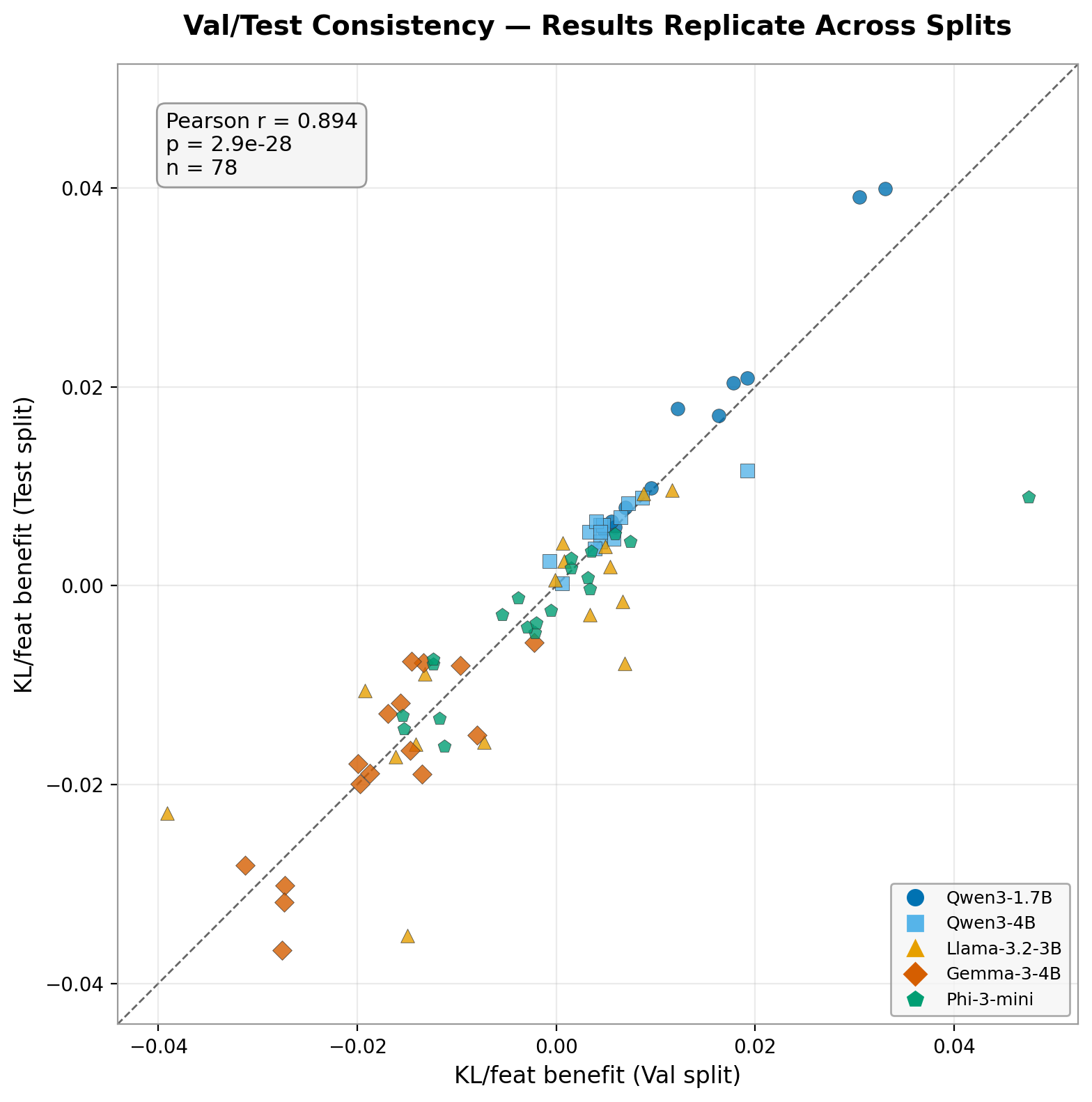}
  \caption{Val vs.\ test KL/feature benefit for all 78 observations
    (descriptive Pearson $r = 0.894$; layer observations are nested
    within models and should not be treated as independent samples).}
  \label{fig:val-test}
\end{figure*}

Table~\ref{tab:ci} reports 95\% Wilson score confidence intervals for
GI-SAE win rates. The Wilson intervals support qualitative separation
between the convergent and saturated profiles: the convergent CI lower
bound (64\%) does not overlap with the saturated CI upper bound (27\%).

\begin{table}[h]
  \caption{GI-SAE win rates with 95\% Wilson confidence intervals
    (test split).}
  \label{tab:ci}
  \centering
  \begin{tabular}{lrrrl}
    \toprule
    Scope & Wins & $N$ & Rate & 95\% Wilson CI \\
    \midrule
    Qwen3-1.7B & 11 & 11 & 100\% & [74\%, 100\%] \\
    Qwen3-4B & 14 & 14 & 100\% & [78\%, 100\%] \\
    Llama-3.2-3B & 7 & 17 & 41\% & [22\%, 64\%] \\
    Gemma-3-4B & 0 & 16 & 0\% & [0\%, 19\%] \\
    Phi-3-mini & 7 & 20 & 35\% & [18\%, 57\%] \\
    \midrule
    Low-sharing & 18 & 37 & 49\% & [33\%, 64\%] \\
    Convergent & 20 & 24 & 83\% & [64\%, 93\%] \\
    Saturated & 1 & 17 & 6\% & [1\%, 27\%] \\
    \midrule
    Overall & 39 & 78 & 50\% & [39\%, 61\%] \\
    \bottomrule
  \end{tabular}
\end{table}

Under a layer-level Bernoulli summary, the Qwen intervals lie above
50\% (lower bounds 74\% and 78\%), while Gemma lies below 50\% (upper
bound 19\%). Because layers within a model are correlated, we interpret
these intervals descriptively. The low-sharing CI [33\%, 64\%]
straddles 50\%, consistent with the characterization that GI-SAE helps
only sometimes in this profile.

\section{Qualitative Feature Audit}
\label{app:audit}

To verify that SAE features encode meaningful cross-language discourse
and answer-boundary structure rather than arbitrary sparse coordinates,
we audit the most
consistently shared features from four representative (model, layer)
pairs, one per sharing profile. For each pair, we encode test-split
activations through the SAE, identify features active in $\geq$4 of 6
languages at the pre-answer token, rank by consistency
($n_\text{problems} \times \text{mean activation}$), and map
peak-activation positions back to text snippets.

\begin{table}[h]
  \caption{Cross-language shared feature audit. Bracketed tokens mark
    peak activation positions. All features fire on semantically
    equivalent content across 4--6 languages.}
  \label{tab:audit}
  \centering
  \scriptsize
  \begin{tabular}{llrrlp{7cm}}
    \toprule
    Model & Layer & Feature & Langs & Category & Representative snippets \\
    \midrule
    \multirow{3}{*}{Qwen-4B} & \multirow{3}{*}{L29}
      & F9662 & 5/6 & Boundary
      & EN: ``should be 29\textbf{[.]}''
        / DE: ``\$29.00\textbf{[.]}''
        / ZH: ``29\textit{(yuan)}\textbf{[.]}'' \\
      & & F8519 & 5/6 & Boundary
      & EN: ``answer should be 29\textbf{[.]}''
        / DE: ``correct total\textbf{[.]}'' \\
      & & F3596 & 5/6 & Boundary
      & EN: ``that seems right\textbf{[.]}''
        / FR: ``that's correct\textbf{[.]}'' \\
    \midrule
    \multirow{2}{*}{Gemma-4B} & \multirow{2}{*}{L21}
      & F3411 & 6/6 & Boundary
      & EN: ``final answer is 30\textbf{[.]}''
        / DE: ``gegeben\textbf{[.]}''
        / ZH: ``30\textit{(yuan)}\textbf{[.]}'' \\
      & & F5299 & 4/6 & Boundary
      & EN: ``Seth is 16\textbf{[.]}''
        / FR: ``16 ans\textbf{[.]}'' \\
    \midrule
    \multirow{2}{*}{Llama-3B} & \multirow{2}{*}{L17}
      & F3882 & 4/6 & Transition
      & DE: ``Antwort \textbf{[ist]}''
        / ES: ``respuesta \textbf{[es]}''
        / ZH: ``\textit{huida}\textbf{[:]}'  \\
      & & F6493 & 5/6 & Boundary
      & EN: ``23 jewels\textbf{[.]}''
        / FR: ``23 bijoux\textbf{[.]}''
        / ZH: ``\textit{da'an}\textbf{[:]}'' \\
    \midrule
    \multirow{2}{*}{Phi-mini} & \multirow{2}{*}{L3}
      & F2162 & 6/6 & Step/format
      & EN: ``First, \textbf{[let]}'s find''
        / ES: ``Prim\textbf{[ero]}, calculamos''
        / DE: ``finden \textbf{[wir]} heraus'' \\
      & & F6445 & 6/6 & Format trans.
      & DE: ``JSON-Antwort\textbf{[:]}'
        / FR: ``JSON est donc\textbf{[:]}''
        / RU: ``JSON\textbf{[:]}'' \\
    \bottomrule
  \end{tabular}
\end{table}

Three feature categories emerge:
\begin{itemize}
  \item \textbf{Answer-boundary markers} (F9662, F8519, F3596, F3411,
    F5299, F6493): fire on punctuation (periods, colons) at the
    reasoning$\to$answer transition. Language-invariant because final
    answers are always numeric.
  \item \textbf{Answer-transition phrases} (F3882): fire on the
    verb or copula introducing the answer (``is''/``ist''/``es''/``:''~[zh]).
    Captures syntactic role, not surface form.
  \item \textbf{Step/format markers} (F2162, F6445): fire on tokens
    initiating reasoning steps or format transitions. Active even at L3,
    suggesting early detection of discourse structure.
\end{itemize}

\section{Patching Convention Sensitivity}
\label{app:norm-sensitivity}

The main sweep uses native-scale patching, while SAE training and
geometric evaluation use z-scored activations. To test whether the
causal conclusions depend on this convention, we rerun patching on
representative layers using normalized-space patching: residuals are
normalized before SAE encoding, patched in normalized feature space,
decoded, and unnormalized before injection into the residual stream.

\begin{table}[h]
  \caption{KL/feature under native-scale and normalized-space patching
    on representative layers. Lower is better.     Gemma L21 is omitted because this diagnostic script produced no
    valid normalized-space interventions for that layer; we do not
    interpret this as evidence about Gemma's sharing profile.
    The full main sweep reports Gemma under the native-scale convention.}
  \label{tab:norm-sensitivity}
  \centering
  \scriptsize
  \begin{tabular}{llrrrrrl}
    \toprule
    Model & Layer & BL raw & GI raw & BL norm & GI norm & Stable? \\
    \midrule
    Qwen-4B & L29 & 0.0087 & 0.0044 & 0.0539 & 0.0040 & Yes \\
    Qwen-1.7B & L24 & 0.0081 & 0.0026 & 0.0739 & 0.0020 & Yes \\
    Llama & L17 & 0.0147 & 0.0500 & 0.0139 & 0.0076 & No \\
    Llama & L25 & 0.0093 & 0.0049 & 0.0202 & 0.0076 & Yes \\
    Phi & L3 & 0.0060 & 0.0009 & 0.0117 & 0.0438 & No \\
    Phi & L15 & 0.0034 & 0.0193 & 0.0130 & 0.0014 & No \\
    \bottomrule
  \end{tabular}
\end{table}

The two Qwen convergent layers remain GI-SAE wins under both
conventions, and the aggregate win rate across valid layers is
unchanged at 4/6 under both conventions. However, normalized-space
patching changes absolute shared-feature counts and flips several
individual low/mixed-profile layers. We therefore interpret the
Qwen/convergent conclusion as robust, but treat exact layer-level
verdicts in low/mixed profiles as convention-sensitive.
% ---- end parts/appendix/appendix ----

\end{document}